\documentclass[letterpaper, 10pt, conference]{ieeeconf}      

\IEEEoverridecommandlockouts                              

\usepackage{amssymb}
\usepackage{graphicx}
\usepackage{color}
\usepackage{wrapfig}
\usepackage{color}
\usepackage{amsmath}
\usepackage{bm}
\usepackage{url}
\usepackage{cite}
\usepackage[subrefformat=parens]{subcaption}
\usepackage{subcaption}
\usepackage{multirow}
\usepackage{multicol}
\newcommand{\argmin}{\mathop{\rm arg~min}\limits}
\usepackage{algorithm}
\usepackage{algpseudocode}
\usepackage{makecell}
\usepackage{booktabs}

\newcommand{\OpenComputer}{{\it Open-Computer\,}}
\newcommand{\BookShelf}{{\it Book-Shelf\,\,}}

\title{\LARGE \bf
Trajectory Optimization for Unknown Constrained Systems using Reinforcement Learning
}

\author{Kei Ota$^{1}$, Devesh K. Jha$^{2}$, Tomohiro Oiki$^{1}$, Mamoru Miura$^{1}$, Takashi Nammoto$^{1}$,\\Daniel Nikovski$^{2}$ and Toshisada Mariyama$^{1}$
\thanks{$^{1}$Kei Ota, Tomohiro Oiki, Mamoru Miura, Takashi Nammoto, and Toshisada Mariyama
	are with Information Technology R\&D Center, Mitsubishi Electric Corporation, Japan.
		{\tt\small Ota.Kei, Oiki.Tomohiro, Nammoto.Takashi, Mariyama.Toshisada@ds.MitsubishiElectric.co.jp}}%
\thanks{$^{2}$Devesh K. Jha and Daniel Nikovski are with Mitsubishi Electric Research Labs, Cambridge, MA, USA.
        {\tt\small jha, nikovski@merl.com}}%
}

\begin{document}
	\maketitle
	\thispagestyle{empty}
	\pagestyle{empty}

\begin{abstract}
	In this paper, we propose a reinforcement learning-based algorithm for trajectory optimization for constrained dynamical systems. This problem is motivated by the fact that for most robotic systems, the dynamics may not always be known. Generating smooth, dynamically feasible trajectories could be difficult for such systems. Using sampling-based algorithms for motion planning may result in trajectories that are prone to undesirable control jumps. However, they can usually provide a good reference trajectory which a model-free reinforcement learning algorithm can then exploit by limiting the search domain and quickly finding a dynamically smooth trajectory. We use this idea to train a reinforcement learning agent to learn a dynamically smooth trajectory in a curriculum learning setting. Furthermore, for generalization, we parameterize the policies with goal locations, so that the agent can be trained for multiple goals simultaneously. We show result in both simulated environments as well as real experiments, for a $6$-DoF manipulator arm operated in position-controlled mode to validate the proposed idea. We compare the proposed ideas against a PID controller which is used to track a designed trajectory in configuration space. Our experiments show that our RL agent trained with a reference path outperformed a model-free PID controller of the type commonly used on many robotic platforms for trajectory tracking.
	
\end{abstract}
\section{Introduction}

\begin{figure}
	\centering
	\includegraphics[width=8cm]{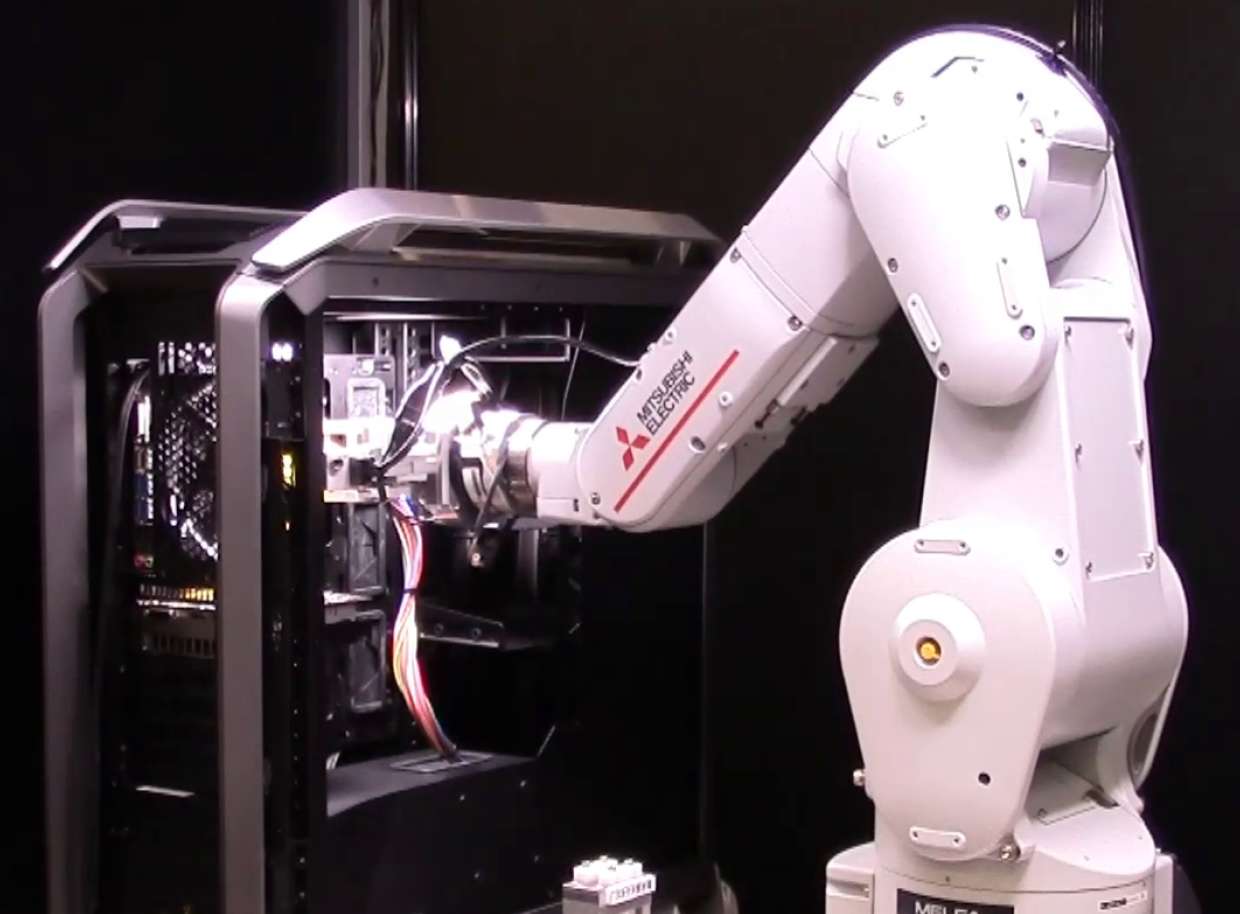}
	\caption{The proposed method is used to compute smooth trajectories for assembling computer parts using a $6$-DoF manipulator arm.}
	\label{fig:manipulator-arm-real-example1}
	\vspace{-4mm}
\end{figure}

In this paper, we present a trajectory optimization algorithm for robots with unknown dynamics operating under state and control constraints using Reinforcement Learning (RL). Trajectory optimization is a procedure that produces state and control sequences for a dynamical system under relevant constraints for the system. Most of the state-of-the-art motion planning algorithms generate a plan in the configuration space of the robot, which is then followed by the robot using a trajectory tracking controller~\cite{lavalle2006planning}. With unknown robot dynamics, the use of traditional PID controllers for trajectory tracking is commonplace. Most of the planning algorithms ignore the dynamics of the robot and hence return a trajectory which is foten unsuitable for a lot of target applications~\cite{lavalle2006planning}. For example, many industrial robots can achieve very high accelerations and torques that can potentially damage the robot as well as the components being manipulated. As an example, in Figure~\ref{fig:manipulator-arm-real-example1} we show a manipulator arm trying to assemble parts of a computer. For such applications, the robot is trying to assemble delicate parts connected with wires with fixed length that can easily be broken if high torques are applied. Furthermore, many users deploying such robots have access only to a high-fidelity simulator for the system and its kinematic equations, but not to the true robot dynamics. As a result, a lot of existing results from model-based trajectory optimization cannot be used. Consequently, for many applications, a lot of time is spent in manually designing trajectory tracking controllers. However, high-fidelity simulators provide a good resource that can be used for training an RL agent in the simulated environment, to be used on the real system later.

A model-free RL agent~\cite{sutton2018reinforcement} can learn feasible trajectories under dynamic constraints; however, it requires a lot of time for convergence to good solutions. In light of this, we propose to train an RL agent to generate dynamically feasible trajectories, where a sampling-based motion planning algorithm is used to guide the RL agent for faster learning and convergence. We use a RRT-based~\cite{lavalle2006planning} trajectory to guide an RL agent to learn dynamically-feasible trajectories in the presence of state and input constraints. The RL agent is trained in a curriculum learning setting for faster convergence, and we show that we can achieve good generalization when conditioning the policy on the goal. The proposed algorithm is tested in several simulated environments, as well as with a real robot, by transferring the learned policy from the simulator to the robot. In Figure~\ref{fig:manipulator-arm-real-example1}, we show a manipulator arm assembling parts in an environment  cluttered with obstacles (a desktop computer) using the proposed algorithm. 

\textbf{Contributions.} Our paper has the following contributions.
\begin{enumerate}
	\item We propose an RL-based algorithm to efficiently generate control-smooth trajectories for unknown constrained dynamical systems using a goal-directed reference trajectory.
	\item The proposed algorithm can generalize to local perturbations in goal position in the presence of obstacles.
	\item We demonstrate the proposed algorithm in both simulation and experimental environments, where the proposed method outperforms a baseline method where a trajectory generated by RRT is tracked by a fine-tuned PID controller. 
\end{enumerate}

\section{Related Work}

Reinforcement learning has recently made huge advances, based on the success of deep learning techniques. Recent RL algorithms have achieved very impressive performance in learning in computer games~\cite{silver2017, mnih2015}, robotics~\cite{levine2016}, etc. Broadly, RL algorithms can be classified as model-based or model-free~\cite{sutton2018reinforcement, KBP13}. Model-based algorithms can achieve good sample complexity and generalization, but are generally known to be harder to train for nonlinear dynamical systems. Model-free algorithms, on the other hand, can achieve good asymptotic performance, but suffer from high sample complexity. A lot of recent research has focused on leveraging ideas from control and optimization theory for faster learning~\cite{schulman2015trust,levine2013guided,levine2014learning}. In a lot of robotics applications, it is generally advantageous to initialize RL agents with demonstrations which can provide them with an initial reference solution~\cite{nair2018overcoming}. Learning policies from reference trajectories has been studied in \cite{pastor2009learning,mulling2013learning,thomas2018learning}. Motivated by this idea, our work mainly focuses on using reference trajectories that can be provided by off-the-shelf planning algorithms to speed up learning for our RL agent. The planning algorithm serves as a demonstrator for the learning algorithm. The closest work similar to ours is~\cite{thomas2018learning}. In~\cite{thomas2018learning}, a model-based RL agent is learned using a trajectory-centric RL (Guided Policy Search) approach to learn a trajectory-tracking controller for a trajectory provided by RRT~~\cite{lavalle2006planning}. However, using a model-based RL in constrained state and control settings could be difficult, because it is not clear how the underlying trajectory optimization algorithm~\cite{tassa2012synthesis} can account for arbitrary state constraints for manipulator-like systems. 


Our combination of RL and reference trajectory tracking can be seen as a form of reward shaping \cite{ng1999policy}.
Reward shaping speeds up learning by creating a more informative reward signal. However, designing shaping rewards requires significant non-trivial reward engineering, and may also alter the optimal solution.
To alleviate this problem, automatic reward shaping has been researched~\cite{konidaris2006autonomous,marthi2007automatic}.




\section{Background\label{sec:background}}

We consider the standard RL setting that consists of an agent interacting with a stochastic environment.
An environment consists of a set of states $\mathcal{S}$, a set of actions $\mathcal{A}$, a distribution of initial states $p(s_0)$, a reward function $r : \mathcal{S} \times \mathcal{A}\rightarrow \mathbb{R}$, transition probabilities $p(s_{t+1}|s_t, a_t) : \mathcal{S} \times \mathcal{A}\rightarrow \mathcal{S}$, and a discount factor $\gamma \in [0,1]$.

An episode starts with an initial observation $s_0$ sampled from $p(s_0)$.
At each time step $t$, the agent observes an observation $s_t$ and chooses an action $a_t$ according to a policy $\pi(a_t|s_t)$, which is a mapping from observations to actions: $\pi : \mathcal{S} \rightarrow \mathcal{A}$.
Then, the agent obtains a reward $r_t = r(s_t, a_t)$, and the next state $s_{t+1}$ is sampled from $p(s_{t+1}|s_t,a_t)$.
The goal of the agent is to maximize the expected discounted sum of rewards $J = \mathbb{E}_\pi[ \sum_{t=0}^{\infty}\gamma^t r(s_t,a_t) ]$.
The quality of the agent's action $a_t$ when receiving an observation $s_t$ can be measured by a $Q$ function $Q(s_t, a_t) = \mathbb{E}_\pi[ J | s_t, a_t ]$.

DDPG \cite{lillicrap2015continuous} is a model-free Q-learning-based reinforcement learning algorithm for continuous action spaces. It is an extension of the earlier DQN agent~\cite{mnih2015}, using generating distributions over continuous action spaces. In DDPG, we maintain two neural networks: a deterministic policy (called the actor) $\pi$ and a Q function approximator (called the critic) $Q: \mathcal{S} \times \mathcal{A} \rightarrow \mathbb{R}$, parameterized by a set of parameters $\theta^\pi$ and $\theta^Q$.
An actor network deterministically maps observations to actions and tries to maximize $Q(s_t, a_t)$.
DDPG employs a critic neural network to estimate Q by minimizing the Bellman loss:
\begin{equation}
L _ { \text {critic} } = \left( Q \left( s _ { t } , a _ { t } \right) - y_t \right) ^ { 2 },
\end{equation}
where the $1$-step target $y_t$ is calculated using target networks $Q^*$ and $\pi^*$ as
\begin{equation}
y_t = r _ { t } + Q ^ { * } \left( s _ { t + 1 } , \pi ^ { * } \left( s_{t+1} \right) \right).
\label{eq:critic_1_step_target_ddpg}
\end{equation}
Each transition of the agent is stored in a replay buffer, from which mini-batches are sampled to train the networks.
This stabilizes training by removing temporal correlations, and therefore reduces the changes in the distributions the networks are trying to learn. Additionally, a prioritized replay buffer~\cite{DBLP:journals/corr/SchaulQAS15} assigns a priority $p_i$ to each transition, computed as the last temporal difference (TD) error and a small hyper-parameter $\epsilon$.
For more details, see \cite{DBLP:journals/corr/SchaulQAS15}.

However, earlier research has shown that DDPG is prone to overestimating Q-values, and results in sub-optimal policies.
TD3 \cite{DBLP:conf/icml/FujimotoHM18} implements three improvements to address the overestimation resulting from approximation errors.
First, it maintains two independent critic networks, and always employs the minimum Q-value as the optimization target.
Second, it proposes to delay the propagation of weight updates.
Finally, it explicitly increases the smoothness of the Q-function prediction by adding a clipped normal noise to the action to  the target Q-value.
Using these three improvements, we can replace the $1$-step target of the critic defined in \eqref{eq:critic_1_step_target_ddpg} with
\begin{equation}
y_t = r _ { t } + \min _ { i = 1,2 } Q _ { i } ^ { * } \left( s _ { t + 1 } , \pi^* \left( s_{t + 1} \right) + \operatorname { clip } ( \mathcal { N } ( 0 , \sigma ) , - c , c ) \right).
\label{eq:critic_1_step_target_td3}
\end{equation}





\section{Proposed Approach}\label{sec:proposed_approach}
In this section, we present details of the algorithm and some techniques which allow us to train the algorithm efficiently. We train a TD3 agent (see Section~\ref{sec:background}) using a reference trajectory provided by RRT. Furthermore, due to the constrained nature of the problem (presence of obstacles), we use curriculum learning to simplify learning for the TD3 agent. These are explained in detail next, and presented as a psuedo code for clarity in Algorithms~\ref{alg:whole_learning_procedure} and~\ref{alg:env}.

\subsection{Reinforcement Learning with Reference Trajectory\label{subsec:proposed_accel_rl}}
We consider the standard RL problem described in Section \ref{sec:background} with a reference trajectory $z \in \mathcal{Z}$.
We include the information about the reference trajectory into a reward function as $r : \mathcal{S} \times \mathcal{A} \times \mathcal{Z} \rightarrow \mathbb{R}$.
Therefore, the reward function can be written as
\begin{equation}
r(s_t, a_t, z_t) = f(s_t, a_t) + h(s_t, a_t, z_t).
\end{equation}

$f(s_t, a_t)$ is the reward that originates from a pure RL setting, and $h(s_t, a_t, z_t)$ is calculated using the reference trajectory.
The idea is to accelerate the learning process by the additionally defined term $h(s_t, a_t, z_t)$ in the reward function. This term penalizes search too far from the reference trajectory, and thus limits the search space for the agent.


In prior work, an expert trajectory is generally used to define the function $h(s_t, a_t, z_t)$. In contrast, we use a standard RRT algorithm for generating a reference path, because the computational cost for generating a path is much smaller than that of doing RL.
Due to the nature of random-sampling based algorithms, RRT produces a jerky path, and it results in jerky trajectories, because the critic directly optimizes the reward function.
In order to mitigate this problem, we investigated two improvements.

First, we reduce the number of vertices that describe  the trajectory by randomly short-cutting between them, as described in \cite{hauser2010}.
To do so, we randomly pick two vertices, and divide the trajectory that connects the two points with a fixed distance.
Then, for each vertex, we check if it has contact with obstacles or not, and if none of the vertices collides with obstacles, we short-cut the path, i.e. omit the vertices between the selected two vertices.

Second, we replace the reference trajectory in every episode with a path found by the RL agent during training that satisfies that: 
1) the current path reaches the goal without colliding with obstacles;
2) the number of total steps to achieve the goal is the lowest, and
3) the cumulative reward is the highest.

\subsection{Resets to Reference Path\label{subsec:reset_to_ref}}
To overcome the problem of exploration, we reset some training episodes to a reference path with a probability of $p_\text{reset} \in [0,1]$.
Restarts from them makes the agent explore more efficiently, because the reference trajectory is guaranteed to reach the goal.
Prior work \cite{nair2018overcoming} employs expert trajectories and resets to a state in them; however, we do not have such expert trajectories.
Instead, we utilize a reference path that ensures that the goal is reached.
To reset to a reference path, we uniformly sample joint angles $\theta_t$ from the set of reference trajectories, and assign the start angle to the sampled value.



\subsection{Curriculum Learning\label{subsec:curriculum}}
Generally, RL is harder to train if the reward function is sparse, and if an episode is longer. To simplify learning for the agent, we gradually increase the complexity of the problem in a curriculum learning setting. We use curriculum learning in two different settings. First, we train the RL agent to learn a controller close to the reference trajectory without any sparse penalty for collision (i.e., $w_3=0$), with the intuition that state constraints make the problem harder. This in turn provides more dense rewards to reach the goal. Once the agent learns successfully how to learn in the absence of obstacles, we introduce penalty for collision, which we gradually increase, to adapt to the obstacles. Second, to make learning easier, we gradually decrease the goal region for the agent. For our problem, we define an acceptable goal area as $r_{\rm{goal}}$, and we declare success when $\bm{\theta}_{\text{dist}}=\| \bm{\theta}_t- \bm{\theta}_{\rm{goal}}\|_2 \leq r_{\rm{goal}}$. (Note that $\bm{\theta}_t$ is six-dimensional).


The gradual change in penalty for collision $w_3$ forces the critic to fit to noisy targets.
This could lead the actor to converge to a non-optimal local minimum, since the actor learns with gradients computed using the critic network. To avoid this, we store past good experiences in a replay buffer, which is different from the prioritized replay buffer that was described in \ref{sec:background}, and encourages the actor to choose the same action as the past good experiences in any given state.
The additionally prepared buffer stores the past best $K$ episodes as in \cite{oh2018}, in the sense that
1) the RL agent reaches the goal without colliding with any obstacle, and
2) the RL agent gets higher episode rewards.

To imitate from such good experiences, we use the behavioural cloning loss $L_\text{BC}$ which was proposed in~\cite{nair2018overcoming} and is defined as:
\begin{eqnarray}
L_{\text{BC}} = \left\{
\begin{array}
{ll} {\left(\pi\left(s_{i}\right) - a_{i} \right)^{2},} & {\text{if } Q \left(s_{i}, a_{i} \right) > Q \left(s_{i}, \pi \left(s_{i} \right) \right)  } \\
{0} & { \text{ otherwise }}
\label{eq:bc_loss}
\end{array}.
\right.
\end{eqnarray}
After sufficient training, the agent might surpass the performance of the past best experiences and thus $L_{\text{BC}}$ would then become detrimental to the agent's performance.
The Q-filter mitigates this problem by only applying $L_{\text{BC}}$ if the critic judges that the action proposed by the actor is worse than the action of the demonstrator, in out setting, past good experiences.
Based on adding a behavioral cloning loss, the actor loss results in:
\begin{equation}
L_\text{actor} = -L_\text{critic} \left( \pi \left( s_{t} \right) \right) + \lambda _ \text{BC} L_\text{BC},
\end{equation}
where $\lambda_\text{BC} \in \mathbb{R}^+$ is a hyperparameter for actor to balance learning from critic or past good experiences.

The whole learning procedure is provided as psuedo code in Algorithm~\ref{alg:whole_learning_procedure} and Algorithm~\ref{alg:env}. These algorithms are implemented in a curriculum setting, as described above.

\subsection{Goal Parameterization of Policy}\label{subsec:goal_parameterization}	
To achieve generalization to perturbations in the target state for the agent, we parameterize the policy of the agent on the goal. The idea is that a goal-parameterized policy represented by a network with enough capacity should be able to generalize to perturbations in the goal location. This is a very desirable property to have in the final policy, because robots are often expected to adapt to some local perturbations in the target state. We assume that the target state $g$ is sampled from the set $\mathcal{G}$. We train a single network to maximize the expected discounted reward over multiple goal states. The learning problem is to optimize the following expected discounted reward:
\begin{equation*}
	J(\mathbf{\pi})=\sum\limits_g J(\pi_g) = \sum\limits_{g}\mathbb{E}_{\pi_g}\Big[\sum\limits_{t=0}^{\infty} \gamma^t r_g(s_t, a_t)\Big]
\end{equation*}
across all goals $g\sim \mathcal{G}$. The reward $r_g$ is now conditioned on the goal to reflect the fact that rewards depend on the particular goal (or task). This is achieved by increasing the capacity of the network by adding additional input units to the network. In the simplest setting, we achieve this by simply padding extra inputs to the network that contain the goal information. 

\begin{algorithm}
	\caption{Learning procedure}\label{alg:whole_learning_procedure}
	\begin{algorithmic}[1]
		\State Initialize $\tau'$ \Comment{\textbf{Buffer for reference trajectory}}
		\State Initialize $R_A \leftarrow \{\}$ \Comment{\textbf{Replay buffer for RL}}
		\State Initialize $R_K \leftarrow \{\}$ \Comment{\textbf{Replay buffer for top-$K$ episodes}}
		\State Compute a reference trajectory $\tau$ using RRT
		\State Smooth out $\tau$ by short-cutting \Comment{\textbf{Reference trajectory input to RL}}
		\While{Termination condition is False}
		\State $t_{\text{step}} \leftarrow 0$, $\tau'\leftarrow \{\}$
		\State Initialize $s_t$ with initial state defined in each task
		\State Replace the initial state with an uniformly sampled state from $\tau$ with probability $p_\text{reset}$
		\While{$t_{\text{step}}<\text{episode length}$ or $\bm{\theta}_{\text{dist}} \leq r_{\rm{goal}}$}
		\State Sample using Algorithm~\ref{alg:env}
		\EndWhile
		\If{Trajectory reaches the goal}
		\If{$\tau'$ satisfies the update condition}
		\State Update reference trajectory, $\tau \leftarrow \tau'$ 
		\EndIf
		\State Update curriculum learning setting
		\EndIf
		
		\EndWhile
	\end{algorithmic}
\end{algorithm}

\begin{algorithm}
	\caption{Environment sampling}\label{alg:env}
	\begin{algorithmic}[1]
		\State Observe $s_t$ and $r_t$.				
		\State Store data for $t$ into $R_A$: $\{ s_t, r_t, a_t\}$ and $\tau' \leftarrow \tau' \cup \{\theta_t\}$
		\If{$s_t$ concludes an episode}
		\State Perform step of TD3 \Comment{\textbf{Update actor and critic networks weights}}
		\If{Current episode deserves top-$K$ episodes}
		\State Update top-$K$ replay buffer $R_K$
		\EndIf
		\Else
		\State Sample the current policy $a_t \sim \pi(a|s_t)$
		\State Advance the environment by performing $a_t$
		\EndIf
	\end{algorithmic}
\end{algorithm}

\section{System Overview\label{sec:task}}
In this section, we provide relevant details of the simulator and the real system we used in this paper for our experiments.


\subsection{Hardware}
We use a MELFA RV-FR robot, which is an industrial robot that has 6 degrees of freedom \cite{melfa}. The generated trajectories must ensure that joint angles and angular velocities that consist the trajectories are within a known specified range. The robot used in the experiments in the paper is operated in a position control mode where a position command is sent to the robot every $\Delta t=0.0035$ seconds, which comes from the minimum operational time of the industrial robot we used in a real setting. As a result, the control input is the velocity for each joints. We, however, would like to minimize the acceleration (i.e., the derivative of the control signal or the control jumps) during operation. This is a desirable feature for a lot of industrial manipulators where direct torque control is not accessible.

\subsection{Simulator\label{subsec:simulator}}
We utilize a simulator to generate trajectories and then deploy them in a real setting. The simulator is a high-fidelity simulator for the MELFA RV-FR called RT ToolBox3 \cite{rttoolbox}. The baseline controller we use to compare the RL agent in this work is a PID trajectory-tracking controller that can be designed in the simulator given a reference trajectory. For our experiment, this function is our initial baseline, which is described in detail in \ref{sec:experiments}. The simulator has a built-in function for collision checking between the manipulator and obstacles present in the environment, and we use the same function for collision detection during planning. However, the proposed algorithm is agnostic to the collision checking method and simulation environment. 

\begin{figure*}
	\begin{multicols}{3}
		\subcaptionbox{\BookShelf environment\label{fig:sim_bookshelf}}
		{\includegraphics[height=52mm]{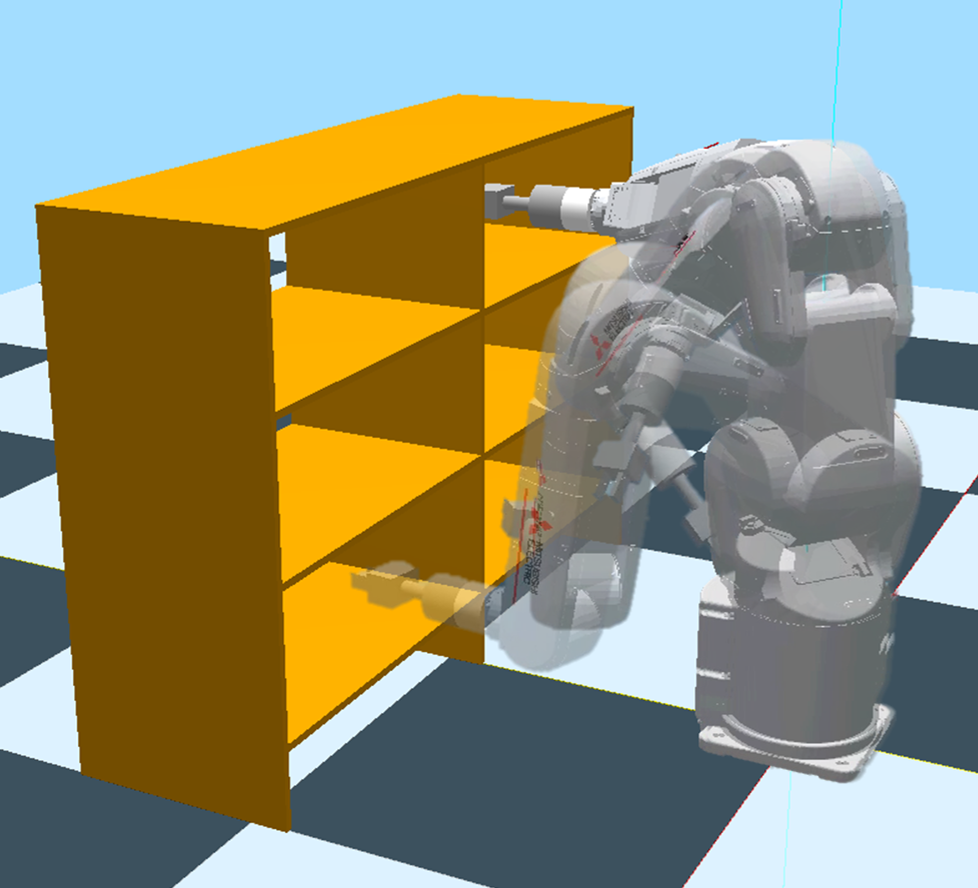}}\par
		\subcaptionbox{\OpenComputer environment\label{fig:sim_pc}}
		{\includegraphics[height=52mm]{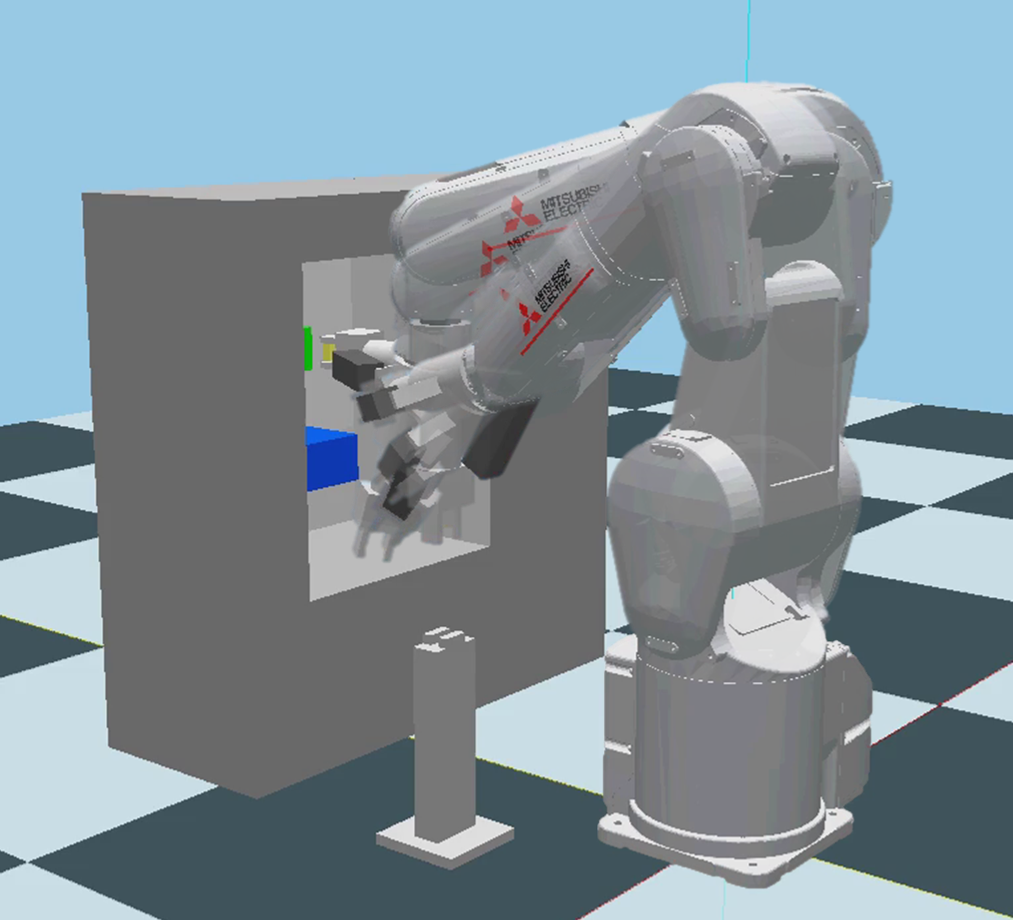}}\par
		\subcaptionbox{\OpenComputer environment in real\label{fig:real_start}}
		{\includegraphics[height=52mm]{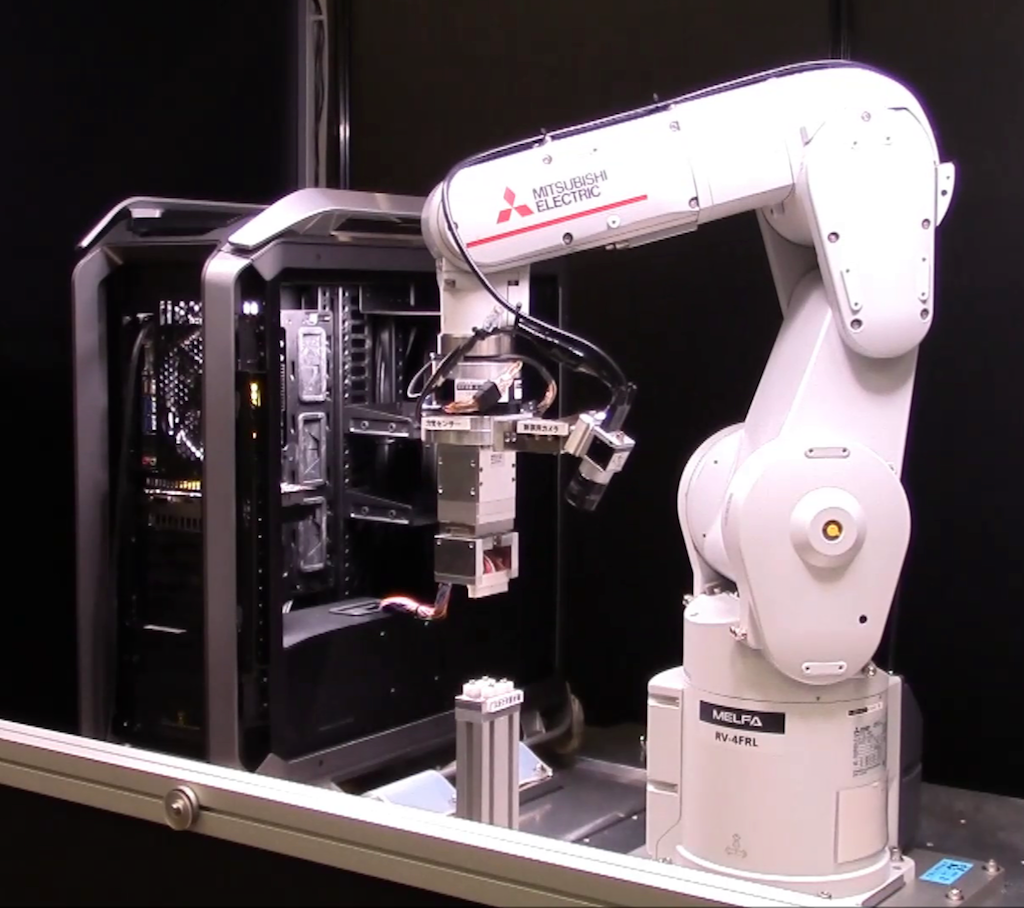}}\par
	\end{multicols}
	\vspace{-2mm}
	\caption{We show the trajectories obtained by the proposed method in the above figures for the three different settings.}
	\vspace{-4mm}
\end{figure*}



\section{Experiments\label{sec:experiments}}
In this section, we will describe several different environments in which we test our proposed algorithm. In particular, we test it in  two environments in simulation--a \textit{Book-shelf} environment (see Figure~\ref{fig:sim_bookshelf}) and an \textit{Open-Computer} environment (see Figure~\ref{fig:sim_pc}). In these environments, the robot is trying to manipulate objects that can be damaged if excessive torque or acceleration is applied. Furthermore, we will show experimental results with a real robot for the \textit{Open-Computer} environment (see Figure~\ref{fig:real_start}). Videos of the learned behavior of the robot could be seen in the supplementary material. 

In our experiments, we try to investigate the following questions:
\begin{enumerate}
	\item Does the combination of a reference trajectory and RL improve the performance of each one of them in isolation?
	\item Does the proposed algorithm generate feasible trajectories in the presence of state and control constraints better than some of the traditional control techniques of trajectory tracking with a reference trajectory?
	\item Does curriculum learning helps the agent learn faster?
\end{enumerate}
In the following text, we answer the above-mentioned questions, and demonstrate that we can generate smooth trajectories and the agent can generalize to unseen goal conditions upon conditioning the policy on goal position. 

\subsection{Environment}
\subsubsection{States}
The states of the system consist of current angles $\bm{\theta}_t$ and angular velocities $\dot{\bm{\theta}}_t$.
Therefore, the state set is represented in $\mathcal{S}\in\mathbb{R}^{12}$.
The initial angles and angular velocities are deterministically reset to 
$\bm{\theta}_0=\bm{\theta}_\text{start}$, and $\dot{\bm{\theta}}_0=\bm{0}$.

\subsubsection{Actions}
The action of the agent $\bm{a}_t$ is the vector of angular velocities $\dot{\bm{\theta}}$ for the next step.
Since we consider a six dimensional configuration space environment, the action set is described as $\mathcal{A}\in\mathbb{R}^6$.
We define a time step described as $\Delta t$.
Therefore, the angles of the next step $\bm{\theta}_{t+1}$ can be calculated as 
\begin{equation}
\bm{\theta}_{t+1}=\bm{\theta}_t + \bm{a}_t \times \Delta t \label{eq:dynamics}.
\end{equation}

\subsubsection{Rewards}
As described in \ref{subsec:proposed_accel_rl}, we add $h(s_t, a_t, z_t)$, which is calculated from the reference trajectory, to the conventional reward term $f(s_t, a_t)$.
First, referring to \cite{vevcerik2017leveraging}, we define the conventional reward term as
\begin{equation}
\begin{split}
f(s_t, a_t) = 
& w_1 d_{\rm goal} + w_2 \mathbb{I}_{\rm goal} + w_3 \mathbb{I}_{\rm collision}\\
& + w_4 \mathbb{I}_{\rm angle limit} + w_5 \| \ddot{\bm{\theta}} \| + w_6,
\label{eq:reward_rl}
\end{split}
\end{equation}
where $d_\text{goal}$ is an Euclidean distance to the goal,
i.e., $d_\text{goal} = \|\bm{\theta}_t - \bm{\theta}_{\rm goal}\|$.
$\mathbb{I}_{\rm goal}$, $\mathbb{I}_{\rm collision}$, and $\mathbb{I}_{\rm angle limit}$ are
indicators of whether the agent reaches the goal, and
whether collision between the agent and the obstacles occurs, and
whether the agent violates the constraint of joint angles respectively.
The fifth term encourages the agent to generate a smoother trajectory, which is essential when operating the real system.
The final term is negative value, so it encourages the agent to reach goal with smaller steps.

Then, we design an additional term by using a reference path as
\begin{equation}
h(s_t, a_t, z_t) = w_7 d_{\rm path} + w_8 n_{\rm progress},
\label{eq:reward_rrt}
\end{equation}
where $d_{\rm path}$ is the distance to the reference path and $n_{\rm progress}$ is the progress along the path.
The first term penalizes search too far from reference path, and second term encourages to go towards goal target angles along with reference path.

\begin{figure}[htpb]
	\centering
	\includegraphics[width=65mm]{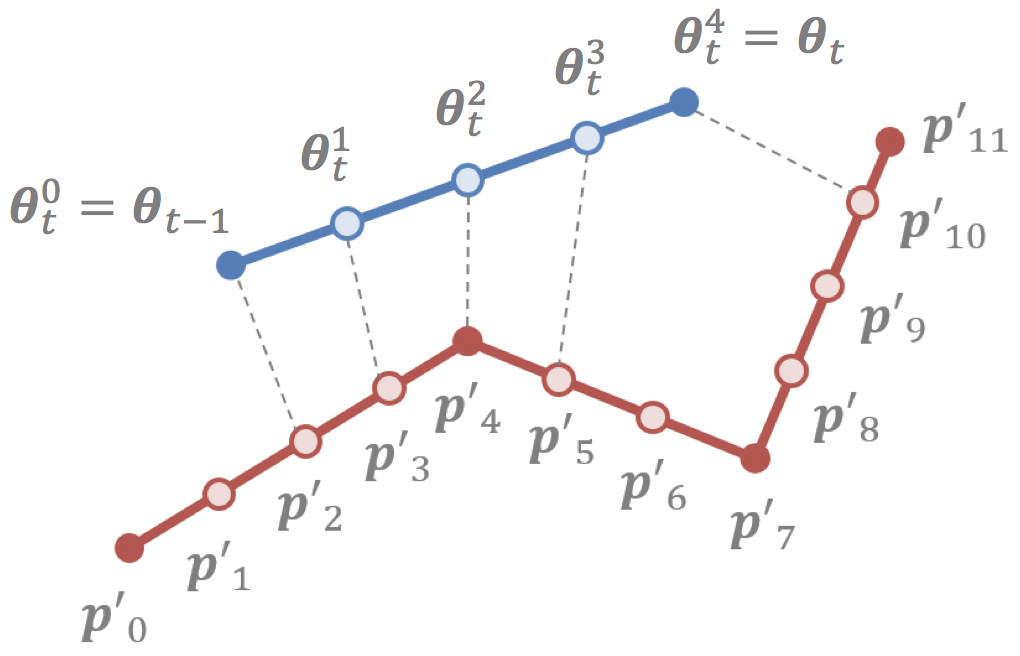}
	\caption{
		Path division for calculating rewards.
		The blue line is the agent's path, and the red line is the reference path.
		$\bm{\theta}_t$ is the joint angles of the agent at time step $t$, and $\bm{p}'_j$ is the index of the divided path.
		The dashed lines indicate the correspondence to the nearest neighbor.}
	\label{fig:path}
\end{figure}

In order to calculate $d_{\rm path}$ and $n_{\rm progress}$, we divide the reference path and agent's path
at regular intervals, as shown in Fig. \ref{fig:path}.
By dividing the path, we obtain the subsampled vertices $\bm{p}'_0, \bm{p}'_1, \cdots, \bm{p}'_{N_p'-1}$ for the reference path,
and $\bm{\theta}^0_t, \bm{\theta}^1_t, \cdots, \bm{\theta}^{N_t-1}_t$ for the agent's path,
where $N_p'$ and $N_t$ are the numbers of vertices in each divided path.
We can then define the distance to the given path as
$d_\text{path} = \max\limits_{i} D(\bm{\theta}_t^i)$,
where $D(\bm{\theta})$ is the distance to the path calculated as $\min\limits_{i}\|\bm{\theta} - \bm{p}'_i\|$.
We can also observe the progress along the path as
$n_{\rm progress} = {\rm NNI}(\bm{\theta}_t) - {\rm NNI}(\bm{\theta}_{t-1})$,
where ${\rm NNI}(\bm{\theta})$ is the vertex index of the nearest neighbor to $\bm{\theta}$,
i.e., ${\rm NNI}(\bm{\theta}) = \argmin_{i}\|\bm{\theta} - \bm{p}'_i\|$.

\begin{figure*}[t]
	\begin{multicols}{3}
		\centering
		\subcaptionbox{\BookShelf Task2\label{fig:episode_return_bookshelf_1}}
		{\includegraphics[width=\linewidth]{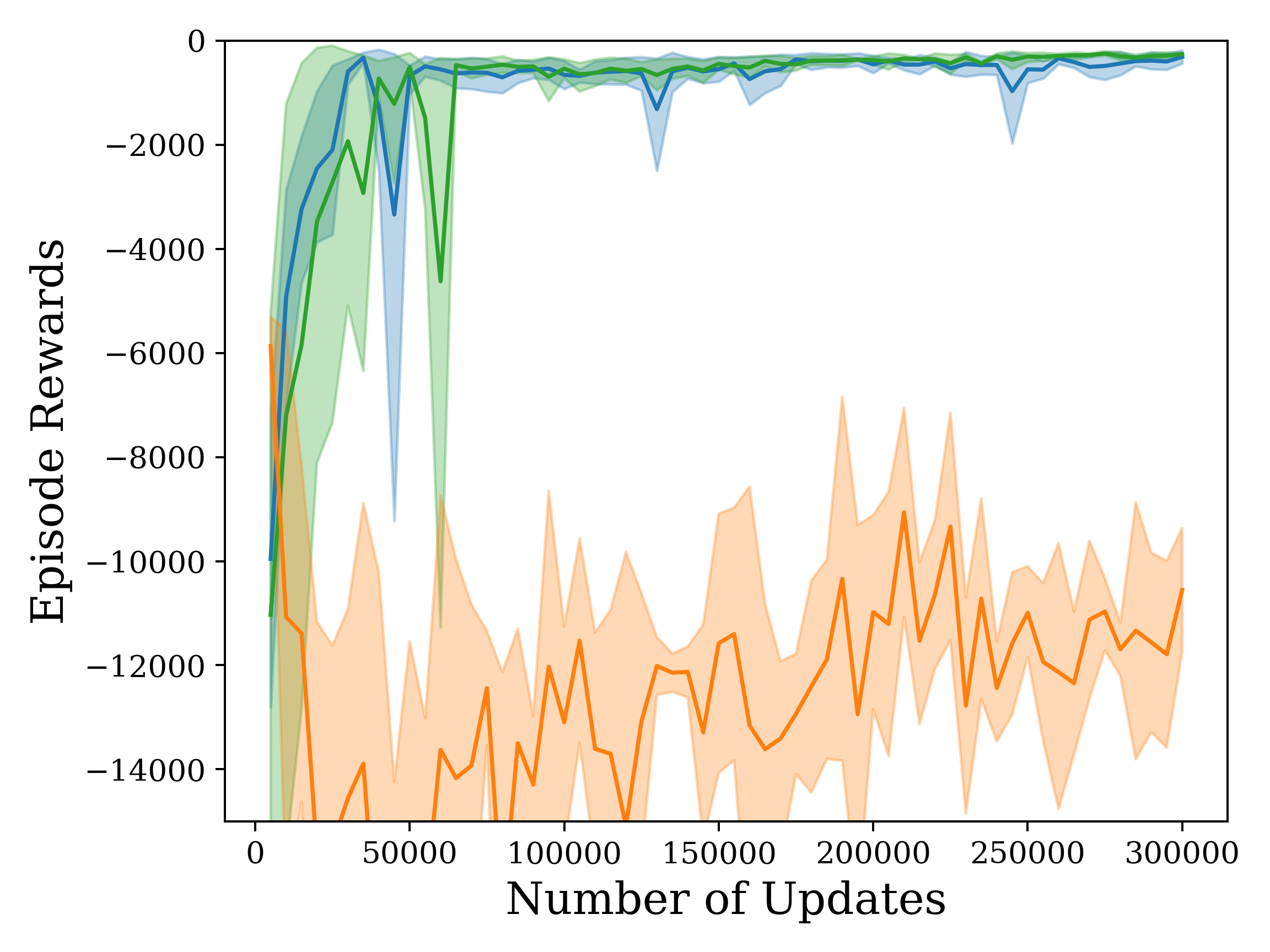}}\par
		\subcaptionbox{\BookShelf Task6\label{fig:episode_return_bookshelf_4}}
		{\includegraphics[width=\linewidth]{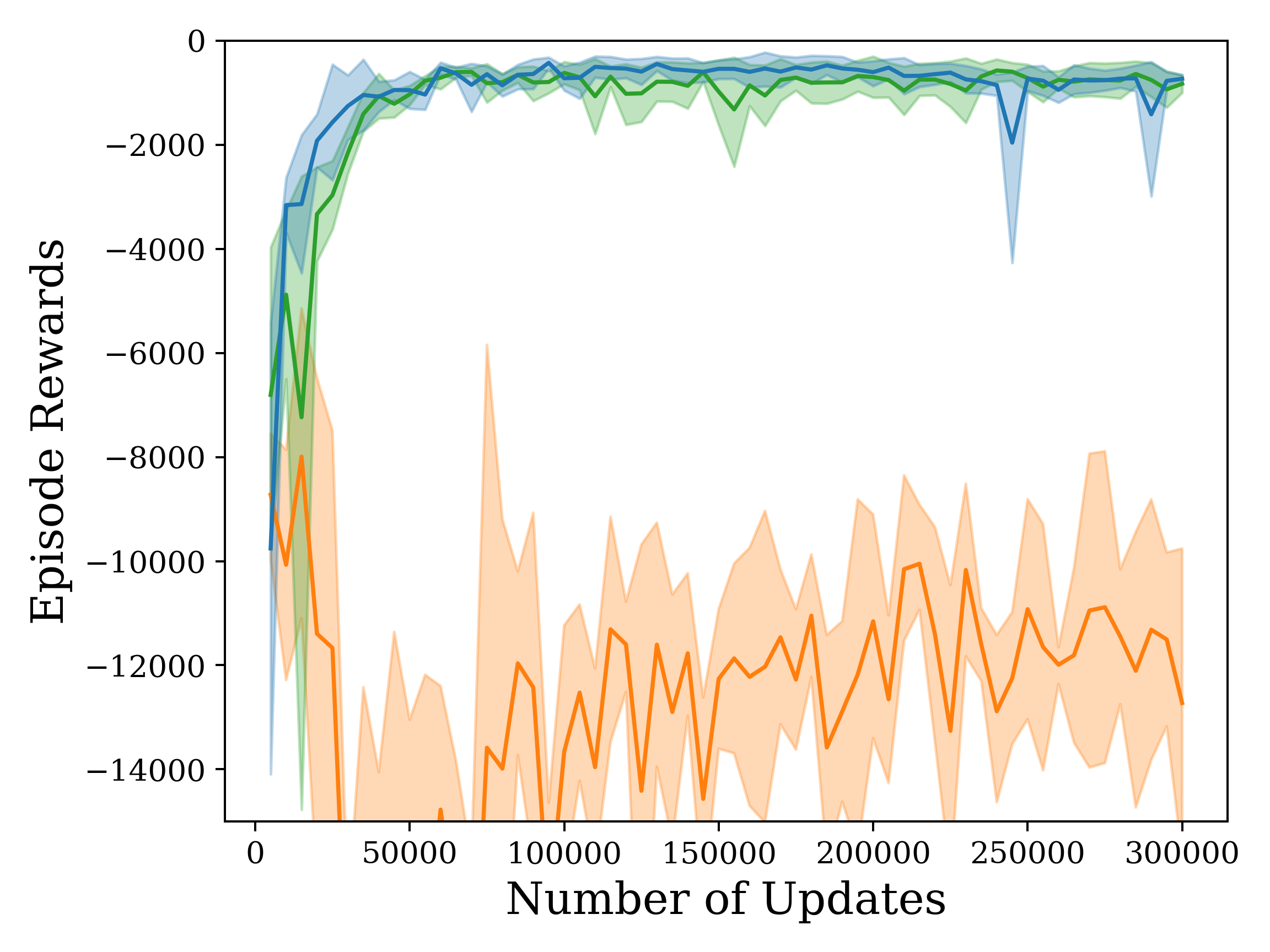}}\par
		\subcaptionbox{\OpenComputer Task\label{fig:episode_return_open_computer}}
		{\includegraphics[width=\linewidth]{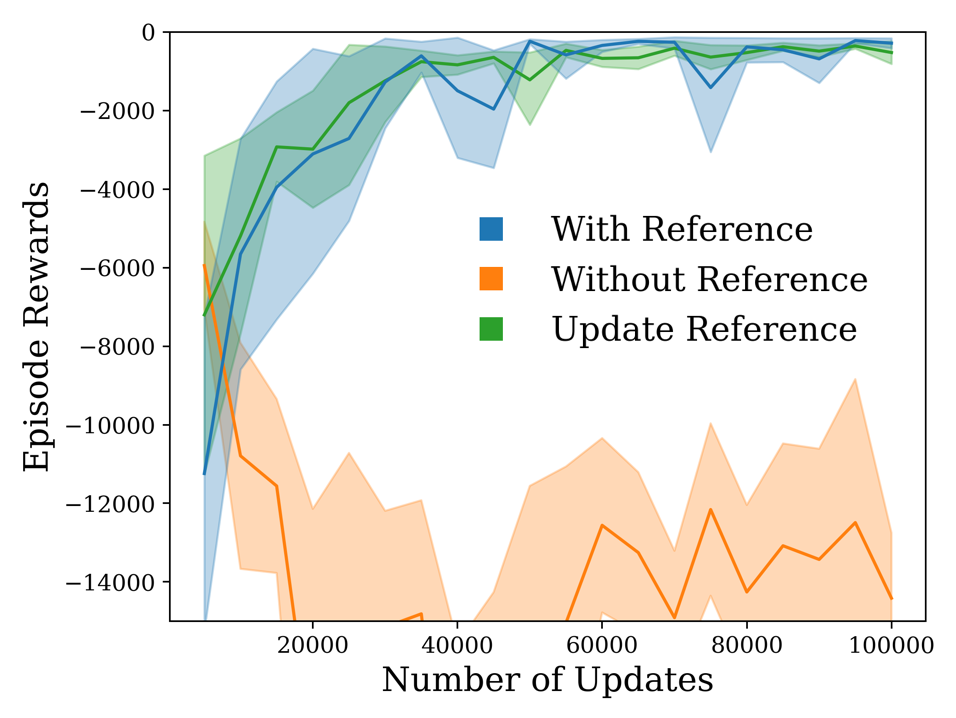}}\par
	\end{multicols}
	\caption{
		A comparison between RL without reference paths, RL with reference paths, and RL with reference paths that are updated in the course of learning. The experiments are conducted over $5$ random seeds.
		The bold line shows the average episode rewards, and the shaded region is one standard deviation from the average. The plot shows the faster and stable learning that we achieve using a reference trajectory which is the updated for self imitation in each episode.}
	\label{fig:results-accelerate-rl}
\end{figure*}

\subsubsection{Termination Condition}
An episode terminates with following two conditions: the joint angles of the agent $\bm{\theta}_t$ are sufficiently close to the goal state as described in \ref{subsec:curriculum}, or the number of steps of an episode is over a specified threshold.

\subsection{\BookShelf Environment\label{subsec:bookshelf_env}}
The \BookShelf environment consists of a two-row, three-stage bookcase, simulating a pick and place task. Each of the cube in the bookshelf is $300$ mm deep, $200$ mm high and $500$ mm wide.
The manipulator starts from an initial pose denoted by $\bm{\theta}_\text{start}$,
and has to reach $6$ different points specified as $\bm{\theta}_\text{BL}, \bm{\theta}_\text{ML}, \bm{\theta}_\text{TL}, \bm{\theta}_\text{BR}, \bm{\theta}_\text{MR}, \bm{\theta}_\text{TR}$, those are center positions for each cube of the bookshelf, defined as 
$\bm{\theta}_\text{start} =$ [0, 8, 131, 0, 41, 180],
$\bm{\theta}_\text{BL} =$ [-52, 59, 106, -141, 78, 170],
$\bm{\theta}_\text{ML} =$ [-52, 28, 111, -134, 60, 152],
$\bm{\theta}_\text{TL} =$ [-52, 13, 95, -111, 42, 117],
$\bm{\theta}_\text{BR} =$ [-128, 59, 106, 141, 78, 190],
$\bm{\theta}_\text{MR} =$ [-128, 28, 111, 134, 60, 208],
$\bm{\theta}_\text{TR} =$ [-128, 13, 95, 111, 42, 243].
We define task $1$ to $6$ as reaching from $\bm{\theta}_\text{start}$ to those angles defined above.


\subsection{\OpenComputer Environment}
The \OpenComputer environment is for simulating a computer assembly, picking up a connector, and inserting it into a socket mounted on a mother board as illustrated in Fig.~\ref{fig:sim_pc}.
The picking part and insertion part is out of our focus, so the simulation starts from just above the connector place with an angle of $\theta_\text{start}$= [-47, -8, 113, 0, 75, -138], and the goal is near the socket, denoted by $\theta_\text{goal}$ = [-90, -1, 138, -180, 46, 88].
The real setup shown in Fig.~\ref{fig:real_start} is the same as the above environment, except that the robot is grasping a connector with a harness.
A video of the implementation of the algorithm on the real manipulator is provided in the supplementary materials.

\section{Experimental Results\label{sec:experimental_results}}

This section presents results from experiments designed to answer the questions described in \ref{sec:experiments}.
The baseline that we compared our proposed method with is a combination of a reference path and a PID-based trajectory tracking controller implemented in our simulator, as described in \ref{subsec:simulator}.
Note that the reference path is generated using RRT and is smoothed out by short-cutting, as described in \ref{subsec:proposed_accel_rl}.

\subsection{Accelerating RL by Using Reference Paths\label{subsec:results_accel_rl}}
First, we evaluate the effectiveness of using reference paths to train an RL agent.
We compare three learning methods.
Firstly, we train an RL agent without a reference path by setting $w_7 = w_8 = 0$ in Eq.\eqref{eq:reward_rrt}. 
Secondly, we train with reference paths, and finally, we train with a reference path while it is being updated in every episode if it satisfies the conditions described in \ref{subsec:proposed_accel_rl}.
The evaluation metric is the cumulative episodic reward that an RL agent gets during an episode.
For fair comparison between methods with and without reference paths, we omit reward terms that come from the reference path by setting $w_7 = w_8 = 0$ in Eq.\eqref{eq:reward_rrt}.

Figure \ref{fig:results-accelerate-rl} shows the resulting episodic returns.
It suggests that the use of reference paths improves convergence performance with respect to the training without reference path.
Also, updating the reference path improves the performance more, because the initial reference path is jerky, and that may result in converging to a non-optimal trajectory. Thus, we see that the use of a reference trajectory for training of the RL agent helps in speeding up policy learning.

\subsection{Generating Smoother and Shorter Trajectories using RL\label{subsec:results_traj_quality}}
\begin{figure}[thpb]
	\centering
	\includegraphics[width=\linewidth]{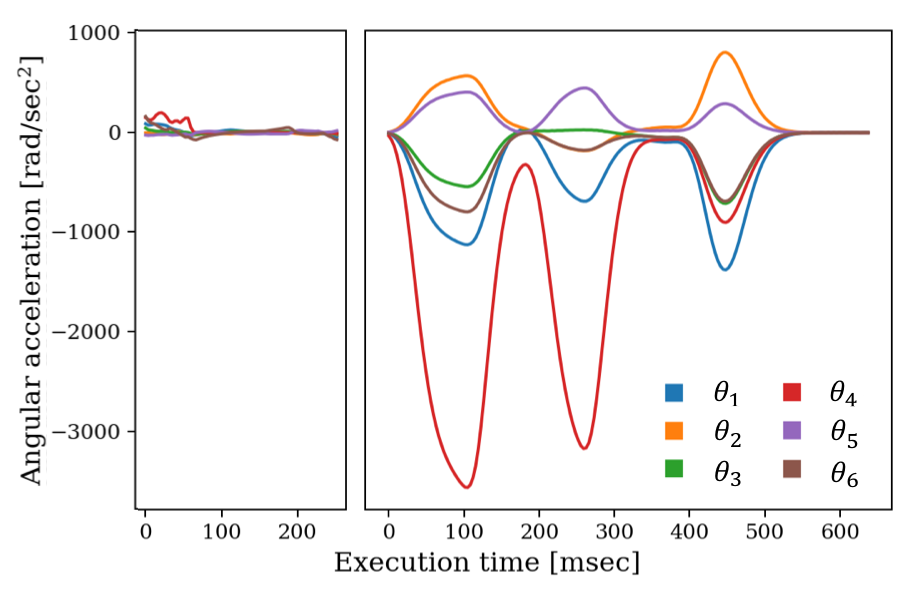}
	\caption{
		Accelerations for \BookShelf environment task1. Left figure is the result of our proposed method and right figure is generated by RRT and PID controller. Note lower is better.}
	\label{fig:angular_accel}
\end{figure}

Next, we compare the quality of the trajectories obtained by the proposed algorithm against the baseline method. We use two metrics to quantify the quality of the trajectories obtained: the time needed to reach the goal, and the magnitude of acceleration. Recall that part of the initial motivation to training the agent this way was to minimize control jumps, and thus generate trajectories with limited acceleration. Table~\ref{table:results_time} shows the time required by the proposed algorithm to reach the goal using the proposed algorithm and compared against the baseline. This clearly demonstrates that the proposed method generates high quality trajectories. Figure~\ref{fig:angular_accel} shows the angular acceleration during a rollout of the proposed method, compared against the baseline method. It shows that the proposed method generates trajectories with much lower acceleration profiles than those generated by the baseline method in all joints, while also minimizing the time taken to reach the goal.  

\begin{table}[htbp]
	\caption{Time [sec] to reach goal.}
	\label{table:results_time}
	\begin{center}
		\begin{tabular}{c|ccccccc} \toprule
			\multirow{2}{*}{Task} & \textit{Open-} & \multicolumn{6}{c}{\BookShelf} \\
			& \textit{Computer} & 1 & 2 & 3 & 4 & 5 & 6 \\ \midrule
			Baseline & 0.82 & 0.56 & 0.65 & 0.62 & 0.80 & 0.75 & 1.28 \\
			Ours & 
			\bf{0.22} & \bf{0.34}& \bf{0.26} & \bf{0.25} & \bf{0.48} & \bf{0.55} & \bf{0.50} \\ \bottomrule
		\end{tabular}
	\end{center}
\end{table}

\subsection{Curriculum Learning}
Next, we investigate how our curriculum learning helps training our RL agent.
We compare our full model with the one without curriculum learning and self imitation for task 1 in the \BookShelf environment.
Note that in TD3, the actor learns to maximize the $Q$ function (critic) parameterized by a neural network.
Therefore, if the estimation of $Q$ function is insufficient, it gives undesirable gradients to the actor and that would result in lower episode rewards.

\begin{figure}[thpb]
	\centering
	\includegraphics[width=0.9\linewidth]{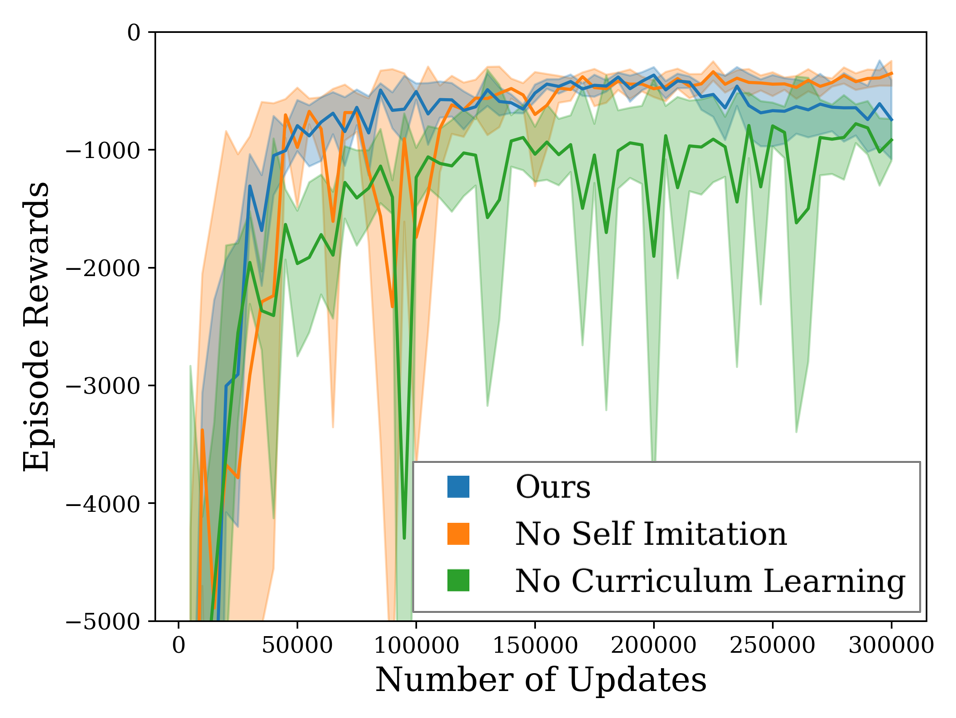}
	\caption{A comparison between our proposed model, removing curriculum learning, and removing self imitation. The experimental conditions are same with Fig.~\ref{fig:results-accelerate-rl}}
	\label{fig:compare_curriculum_learning}
\end{figure}

Fig.~\ref{fig:compare_curriculum_learning} shows a comparison of convergence rates of the agent using different methods.
Without curriculum learning, the agent achieves slower convergence, because the training of the critic is harder due to a huge collision penalty, and it is harder to get positive reward which the agent receives only upon reaching the goal.
Also, training without self imitation results in unstable training, because the critic needs to fit the noisy reward because of changing collision penalty $w_3$ in Eq.~\eqref{eq:reward_rl}.
Our full model is both stable and converges faster, because curriculum learning makes it easier for the critic to fit the $Q$ function, and self imitation mitigates the noisy reward problem by imitating past good experiences.


\subsection{Generalization to Goal Perturbation\label{subsec:goal_generalization}}
Next, we evaluate the generalization of our method with respect to goal perturbation.
As shown in \ref{subsec:goal_parameterization}, we add the goal state to both the actor and critic networks, in the expectation that the method can generalize over goal states.
The target task is task 2 of \BookShelf environment, moving from start angle $\theta_\text{start}$ to $\theta_\text{ML}$ defined in \ref{subsec:bookshelf_env}.
As for targets, we fix a plane in a cube of the \BookShelf and change goal state within a $W \times H$ [mm$^2$] rectangle.
For training, we randomly sample $N$ different positions in the rectangle.
After training, we test the generalization by randomly sampling $50$ positions in the same rectangle used during training, recording whether the robot reached the goal or not.
To exploit the past good experiences, we prepared top-$K$ buffer for each goal, and tried to make the training more stable and improve sample efficiency.

Table \ref{table:goal_param_success_rate} shows the result of the experiments.
It demonstrates that the RL agent successfully generalizes to goal perturbations over a reasonable area even in the presence of state constraints. 

\begin{table}[htbp]
	\caption{Performance on generalization task in simulation}
	\label{table:goal_param_success_rate}
	\begin{center}
		\begin{tabular}{c|cccccc} \toprule
			\multirow{2}{*}{$(W,H,N)$} & \multicolumn{3}{c}{$(100,100,10)$} & \multicolumn{3}{c}{$(200,200,50)$} \\
			& Train & Test & Overall & Train & Test & Overall \\ \midrule
			Number of & \multirow{2}{*}{10/10} & \multirow{2}{*}{49/50} & \multirow{2}{*}{59/60} & \multirow{2}{*}{49/50} & \multirow{2}{*}{45/50} & \multirow{2}{*}{94/100} \\
			successes & & & & & & \\
			Success rate & 1 & 0.98 & 0.98 & 0.98 & 0.90 & 0.94 \\ \bottomrule
		\end{tabular}
	\end{center}
\end{table}





\section{Conclusion}
The research reported in this paper is based on the idea to combine RL with trajectory optimization for unknown systems in the presence of constraints on state, control and control-jumps. This kind of problems is common in robotics, where a manipulator has to be used for tasks in an environment cluttered with obstacles, in a position-control mode. We proposed a method based on RL, for the case when the dynamics are unknown, that generates optimal trajectories in the presence of obstacles and other constraints. For faster learning, we use an off-the-shelf sampling-based algorithm to first generate a reference trajectory which is then used by the RL agent to converge to an optimal solution faster. The proposed method was demonstrated on several simulated environments using a high-fidelity simulator for an industrial-grade manipulator. We compared the learned policy against a baseline controller designed to track the trajectory obtained by smoothing the initial reference trajectory. The proposed algorithm was also tested for generalization to multiple new target states.

In future research, we would like to investigate the proposed algorithm by parameterizing it with the reference trajectory. We expect that as long as we do not change the environment, the agent would learn to produce a better solution respecting all the constraints. 
    %
	\bibliographystyle{IEEEtran}
	\bibliography{IEEEabrv,references}
\section*{Appendix}

\subsection{Curriculum Learning Setting}
As written in \ref{subsec:curriculum}, we used curriculum setting to train our RL agents.
Let $t_\text{goal}$ be the number of goals reached in an experiment.
When $0 \leq t_\text{goal} < 100$, we train RL agents without checking collision with obstacles, and linearly decrease $r_\text{goal}$ from $0.52$ [rad] to $0.26$ [rad].
Then, we linearly increase the value of the collision penalty from $0$ to $-10$ in $100 \leq t_\text{goal} < 200$.

For self imitation, we set $K=5$ for the top-$K$ replay buffer, and do self imitation only when the buffer is filled with $K$ episodes.
For goal generalization experiments in \ref{subsec:goal_generalization}, we set $K=2$ for all different goal settings, and start self imitation when more than 20\% of the top-$K$ replay buffer is stored.

\subsection{Training Details}
Both the actor and critic networks have two hidden layers with 128 and 64 units for each layer.
The hidden layers use the ReLU activation function, and the output layer of the actor uses the tanh activation function, so that an action lies in the range of $[-1,1]$.
We define the maximum step for an episode to be $300$, and the agent randomly resets to a reference path with a probability of $p_\text{reset}=0.3$, as described in \ref{subsec:reset_to_ref}.
We train our TD3 agent for at most one million steps.
Both the actor and the critic perform updating every time an episode finishes, collecting $1,000$ samples, with a minibatch of size $100$ sampled from a prioritized replay buffer. The prioritized replay buffer consists of $10^5$ transitions with fixed $\alpha=0.6$ and $\beta=0.4$.
For the ADAM optimization algorithm \cite{kingma2014adam}, we use learning rates of $0.001$ for both the actor and the critic, and the default values from the TensorFlow framework for the other hyperparameters.
The target networks are also updated every cycle using a decay coefficient of $0.995$.
We use a discount factor of $\gamma = 0.9$.


\end{document}